\documentclass[sigconf]{acmart}

\usepackage{multirow}

\setcopyright{none}
\renewcommand\footnotetextcopyrightpermission[1]{}
\acmDOI{}
\acmISBN{}

\begin{document}

\title{Automated Feedback to Peer Tutors Increases Their Talk Moves and Learner Reasoning but Not the Balance of Talk}

\author{Dorottya~Demszky$^{1}$, Yann~Hicke$^{2}$, Joy~Yun$^{1}$, Mariah~Olson$^{3}$, Max~Karpawich$^{3}$, Anees~Iqbal$^{3}$, Matt~Wu$^{3}$, Jimmy~Huang$^{3}$, Akshay~Ravikumar$^{3}$, Drew~Bent$^{3}$\vspace{+.5em}}
\affiliation{%
  \institution{$^{1}$Stanford University, Stanford, CA, USA \quad $^{2}$Cornell University, Ithaca, NY, USA \quad $^{3}$Schoolhouse.world, USA}
  \country{}\vspace{+.5em}}
\email{ddemszky@stanford.edu, ylh8@cornell.edu, joyyun@stanford.edu, mariahraeolson@gmail.com}
\renewcommand{\shortauthors}{Demszky et al.}

\begin{abstract}
Automated feedback on classroom discourse improves instruction in professional and online teaching contexts, and recent work suggests that delivering similar feedback to learners may strengthen these effects by sharing responsibility for participation. Whether this holds in small-group volunteer tutoring has not been tested. We present results from a randomized controlled trial across two four-week SAT mathematics bootcamps on the Schoolhouse.world platform, involving 1,030 volunteer tutors, 7,443 learners and 6,508 recorded sessions. Tutors were randomly assigned to receive no feedback, automated feedback on their own discourse, or that feedback together with participation feedback delivered to their learners under one of two motivational framings. Feedback to tutors increased eliciting of learner ideas, uptake of learner contributions, and learner reasoning by 8--15\%, and the pattern replicated across both bootcamps. However, tutor talk time, which averaged 91.5\% in the control group, did not change in any condition, and learner-facing feedback did not further change discourse. Interviews with 17 tutors indicate that they adopted the moves the feedback recommended but were limited by learners who did not respond. Interviews with 13 learners suggest why the learner-facing feedback had little effect: it reported a group-level rather than a learner-level statistic, some learners preferred to listen, and learner-to-learner interaction was rare in these sessions. Automated feedback can change how tutors invite learner thinking, but shifting how much learners talk likely requires changes beyond feedback.
\end{abstract}



\maketitle
\pagestyle{plain}

\section{Introduction}

Students have more opportunities to express their thinking when instructors invite their contributions, take them up, and press for reasoning
\cite{chapin2009,franke2001,webb2014}. These practices are difficult to learn without feedback, and the usual source of feedback, an instructional coach or mentor, is expensive and unavailable to most instructors \cite{kraft2018}. The gap is especially wide in volunteer tutoring programs, which rely on novice tutors, often high school students themselves, and provide little individualized coaching.

Automated feedback derived from session recordings offers a scalable alternative. Teachers who used automated feedback on their talk moves reported that it supported reflection \cite{jacobs2022,kelly2024}. Randomized trials have shown that instructors who receive such feedback reduce their talk time and take up more student ideas in online mentoring \cite{demszky2023mpowering}, increase their uptake of student contributions in a large online course \cite{demszky2024uptake}, and ask more focusing
questions in K-12 classrooms \cite{demszky2025questioning}. In each of these studies, the feedback is addressed to the instructor alone.

A recent line of work suggests that delivering feedback to learners as well may be more effective, by sharing responsibility for participation rather than placing it entirely on the instructor. In 1:1 tutoring, a talk ratio meter shown to tutors alone or to both tutors and students increased student talk in both arms, and in the student-facing arm the increase reflected more mathematical reasoning \cite{demszky2024talkmeter}. A pilot on the platform we study found that joint feedback improved practice test scores but also, unexpectedly, increased tutor talk time \cite{yun2024}. \textbf{It is not yet known whether joint tutor-learner feedback helps in small-group instruction}, or when it arrives after the session, separately to each participant, rather than during it in a shared display.

We report a randomized controlled trial across two four-week SAT mathematics bootcamps on Schoolhouse.world, a free online peer tutoring platform. Tutors (1{,}199 tutor assignments; 1{,}030 distinct tutors) were randomly assigned to one of four conditions: no feedback, automated feedback on their own discourse, or that feedback together with participation feedback delivered to their learners, under either a socially-oriented or a goal-oriented framing. Our analysis plan was registered before outcome data were analyzed \anon[(anonymized link)]{(AEARCTR-0014050, \url{https://doi.org/10.1257/rct.14050-1.0})}. We address the
following research questions:

\begin{enumerate}
\item Does automated feedback to tutors change their instructional practice and their learners' outcomes?
\item Does automated feedback to learners, delivered alongside feedback to tutors, change these outcomes further, and does its motivational framing make a difference?
\item How did tutors and learners perceive the feedback, and what were the barriers to acting on it?
\end{enumerate}

We answer the first two questions with quantitative analyses, and the third with interviews with 17 tutors and 13 learners, following the mixed-methods approach used in prior work on automated feedback \cite{demszky2025questioning}.

We find that feedback to tutors increased their eliciting of learner ideas and uptake of learner contributions, as well as learner reasoning, by 8--15\%, and that the pattern replicated across both bootcamps. The balance of talk did not change.  Tutors did about 90\% of the talking in every condition. Interviews indicate that tutors acted on the summary statistics and the generated suggestions, often naming the specific change they made, and that the most common barrier was learners who did not respond. Adding learner-facing feedback did not further change discourse. Automated feedback thus changed how tutors invited learner thinking, and learners responded with more reasoning, but shifting how much learners talk likely requires changes beyond feedback to tutors or learners.

\section{Related Work}

\subsection{Measuring instruction through classroom discourse}

Instruction that encourages students to verbalize and build on each other's thinking supports learning \cite{alexander2008,nystrand1997}. Frameworks for dialogic and accountable talk identify a small set of moves that do this work, including inviting student ideas, revoicing and building on them \cite{oconnor1993}, and pressing for reasoning \cite{herbel2005,chapin2009}. These moves are associated with student outcomes \cite{webb2014}, and they are often the targets of instructional coaching.

Natural language processing has made several of these moves measurable from transcripts at scale. Researchers have developed classifiers for several teacher talk moves using annotated corpora, including teachers' uptake of student ideas \cite{demszky2021uptake}, pressing for reasoning \cite{suresh2022talkmoves}, authentic questions \cite{jensen2021transfer}, and assessing how teachers' attend to student ideas during group discussion \cite{nazaretsky2023attention}. More recently, large language models have been used to code teacher-student discourse at the utterance level against an observation framework \cite[e.g.,][]{xu2024promises,yoo2026dual,vanacore2026moves}. These measures allow an instructor to receive a description of their own discourse without a human observer present, though they capture the presence of a move rather than its quality, and their accuracy depends on how closely a setting resembles the corpora on which they were trained.

\subsection{Automated feedback to instructors}

Randomized trials provide causal estimates of automated feedback on instruction: \citet{demszky2023mpowering} found that mentors who received it decreased their talk time by 5\% and increased their uptake of student ideas by 10\%; \citet{demszky2024uptake} found a 13\% increase in uptake among
instructors in a large online course; and \citet{demszky2025questioning} found a 20\% increase in focusing questions in K-12 classrooms, with no effect on other discourse features. Effects vary across contexts and designs. Teachers can
find automated feedback accurate and reflect on it at length while evidence of uptake in their practice remains limited \cite{kelly2024}, and coaching supported by automated discourse analytics improved tutors' use of four of six targeted moves \cite{sawaya2025}. 
The feedback we study pairs descriptive statistics and detected talk moves from the tutor's own transcript with generated suggestions.

\subsection{Feedback to learners}

Research on feedback literacy treats learners as active interpreters of feedback rather than passive recipients \cite{carless2018}. \citet{hu2024} found that participation analytics made students' collaborative dialogue more equitable in responsiveness and opportunity to participate, and \citet{worden2026llmfeedback} found that generated feedback improved short-term learning for lower-knowledge middle school students. A systematic review of learner-facing dashboards found little effect on achievement, with participation the one domain showing gains \cite{kaliisa2024}, while \citet{ochoa2024} found no improvement in equitable participation and traced this to the varied paths learners take from feedback to self-regulation.


Work on displays of participation suggests a limitation relevant to our study. When groups were shown each member's speaking time, dominant speakers reduced their talk, but quieter members did not increase theirs, and they judged the display less accurate than their more talkative peers did \cite{dimicco2004}. \citet{bachour2010} found a similar asymmetry with an interactive table that visualized participation during collaborative work. \citet{strauss2021} combined a group awareness tool with adaptive prompts to encourage equal participation in online collaboration, found only tentative effects, and reported that students rarely used the support. This indicates that awareness of group imbalance does not in itself increase the amount that quieter speakers talk.

\citet{demszky2024talkmeter} provide the closest experimental evidence in a tutoring setting: an in-session talk ratio meter (Talk Meter) shown to both tutor and student increased student talk ratios by 13--14\% in both treatment arms, and in the student-facing arm the increase reflected more student mathematical reasoning.

\subsection{Discourse feedback in volunteer tutoring}

Tutoring is among the most effective interventions in education, with the largest effects when teachers or paraprofessionals tutor \cite{nickow2024}. Programs have grown to meet demand by recruiting tutors who are not professional educators: volunteer and near-peer models, including high school students tutoring younger peers, are low-cost and scale quickly \cite{kraft2021blueprint}, but rely on tutors with limited training and few opportunities for individualized coaching, which makes automated support particularly relevant.

\citet{wang2024copilot} gave novice tutors real-time, expert-like suggestions during sessions and found that students of supported tutors were four percentage points more likely to master topics, with larger gains for students of
lower-rated tutors. That study and the Talk Meter deliver support during the session; we study feedback delivered after the session, which is cheaper to produce and potentially less distracting. This study thus extends the evidence on automated discourse feedback to three conditions that matter for tutoring programs at scale: feedback after the session, small-group rather than 1:1 instruction, and volunteer peer tutors, most of them high school students. Conducting the trial across two consecutive bootcamps also lets us observe whether the results replicate.

\section{Study Background}

We conducted this study in partnership with Schoolhouse.world, a free online peer-to-peer tutoring platform founded in 2020. Volunteer tutors, most of whom are themselves high school students, lead small group sessions for learners across more than 180 countries. Anyone with an SAT mathematics score of 700 or above may apply to tutor, and all tutors complete asynchronous training on the platform's instructional rubric together with a live onboarding session before their first bootcamp.

The platform offers four-week SAT bootcamps consisting of eight 75-minute sessions, each led by a volunteer tutor with a maximum of ten learners. Sessions take place over Zoom and are recorded. Tutors follow a curriculum covering specific SAT topics each week, and each session is divided between practice tests, review of problems, and tutor-selected exercises. Learners participate both by speaking and by typing in the Zoom chat. Learners join a bootcamp by selecting their subject, score range and preferred times, and then choose among the available sections; enrolling in a section commits them to all eight sessions with that tutor. Learners occasionally switch sections, for instance when their schedule changes or their tutor does not show up. In our data, 0.8\% of learners in the first bootcamp and 2.3\% in the second attended sessions with more than one tutor, and 0.5\% and 1.5\% attended sessions in more than one condition.

We study two consecutive bootcamps in Spring 2024, each preparing learners for a specific SAT administration. The May bootcamp ran in April 2024 to prepare for the May SAT test, and the June bootcamp ran in May 2024 to prepare for the June SAT test; we refer to them as the first and second bootcamp. A pilot randomized controlled trial during the March 2024 bootcamp informed the design of the feedback used here \cite{yun2024}. The present data were used once before, for an offline contextual bandit policy evaluation that does not estimate the trial's treatment effects \cite{yun2025bandits}.

Our study was approved by the Institutional Review Board of \anon[XXX]{Stanford University}. Tutors and learners consented to recording under the platform's terms of use. Data were de-identified and shared with university researchers under a data use agreement.

\section{Randomized Controlled Trial}

\subsection{Experimental conditions}

Before the first session, tutors were randomly assigned to one of four conditions. One third (n=399) were assigned to a control condition and received no feedback. One third (n=401) were assigned to
\textsc{TutorFeedback} and received automated feedback on their own discourse after each session. The remaining third was split evenly between \textsc{Tutor+GoalLearner} (n=199) and \textsc{Tutor+SocialLearner} (n=200); in both, tutors received the same feedback and their learners additionally received feedback on their own participation, framed either in terms of the learner's own goals or in terms of benefit to the group.

Randomization was at the tutor level, so all learners assigned to a tutor experienced the same condition, and a tutor running more than one section in a bootcamp was assigned to a single condition across them. Arms were balanced on the bootcamp's SAT score band and on section size. Tutors taking part in a separate platform study were excluded. Assignment was carried out separately for each bootcamp: of the 1{,}030 distinct tutors in the study, 169 taught in both and were randomized again for the second, 124 of them into a different condition. Some tutors had also received feedback in the March pilot (see Section~\ref{ssec:robustness} for estimates that exclude control tutors with prior exposure). Before the first bootcamp, tutors assigned to a treatment condition received an email from the platform explaining that they would receive feedback; tutors in the second bootcamp were not notified in advance.

\begin{figure*}[t]
\centering
\begin{minipage}[t]{0.50\linewidth}\vspace{0pt}\centering
\includegraphics[width=\linewidth]{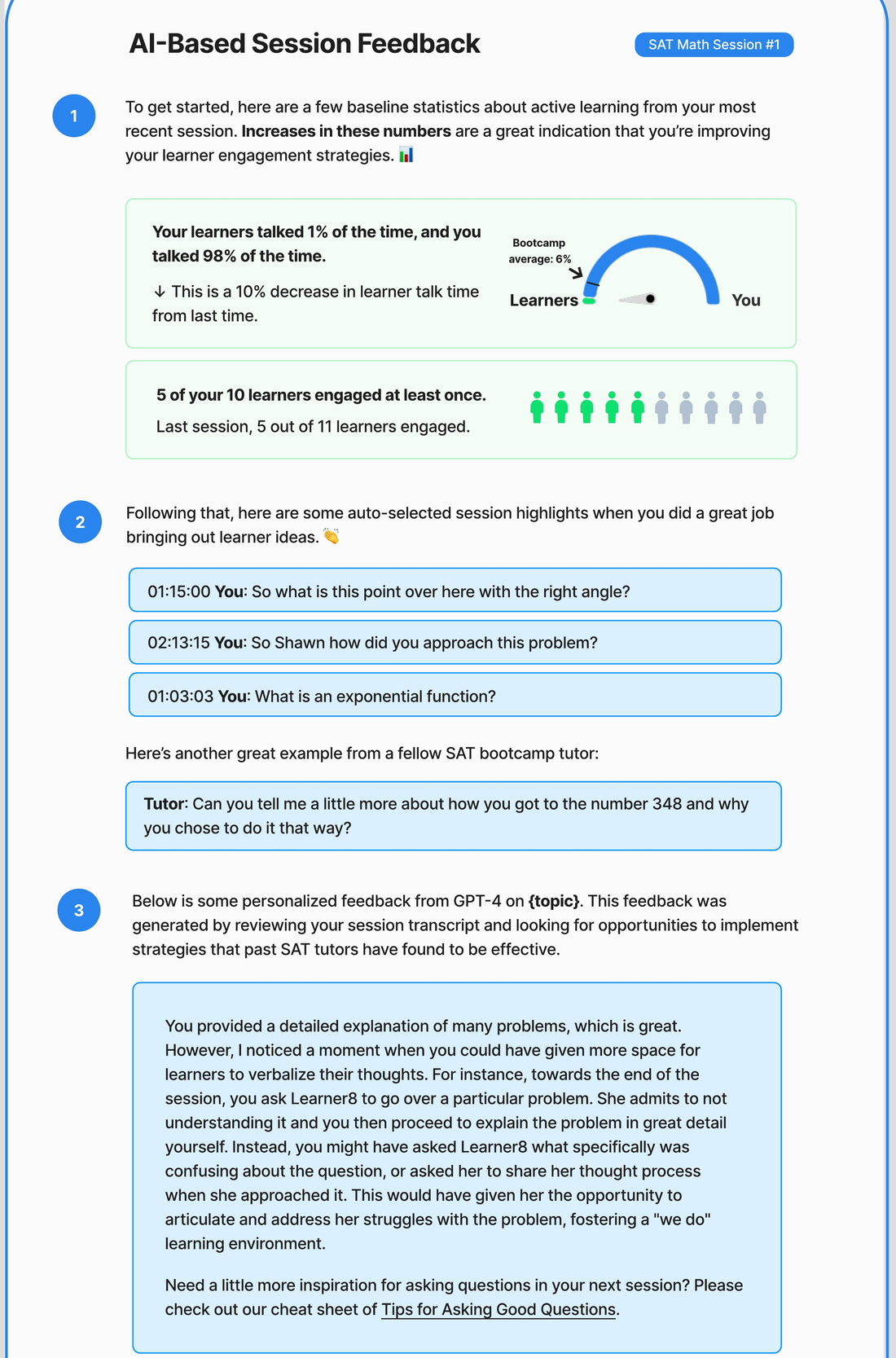}
\Description{Mockup of the tutor feedback modal with summary statistics, highlighted transcript moments and generated suggestions.}\\[3pt]
{\small (a) Tutor feedback, components 1--3}
\end{minipage}\hfill
\begin{minipage}[t]{0.44\linewidth}\vspace{0pt}\centering
\includegraphics[width=\linewidth]{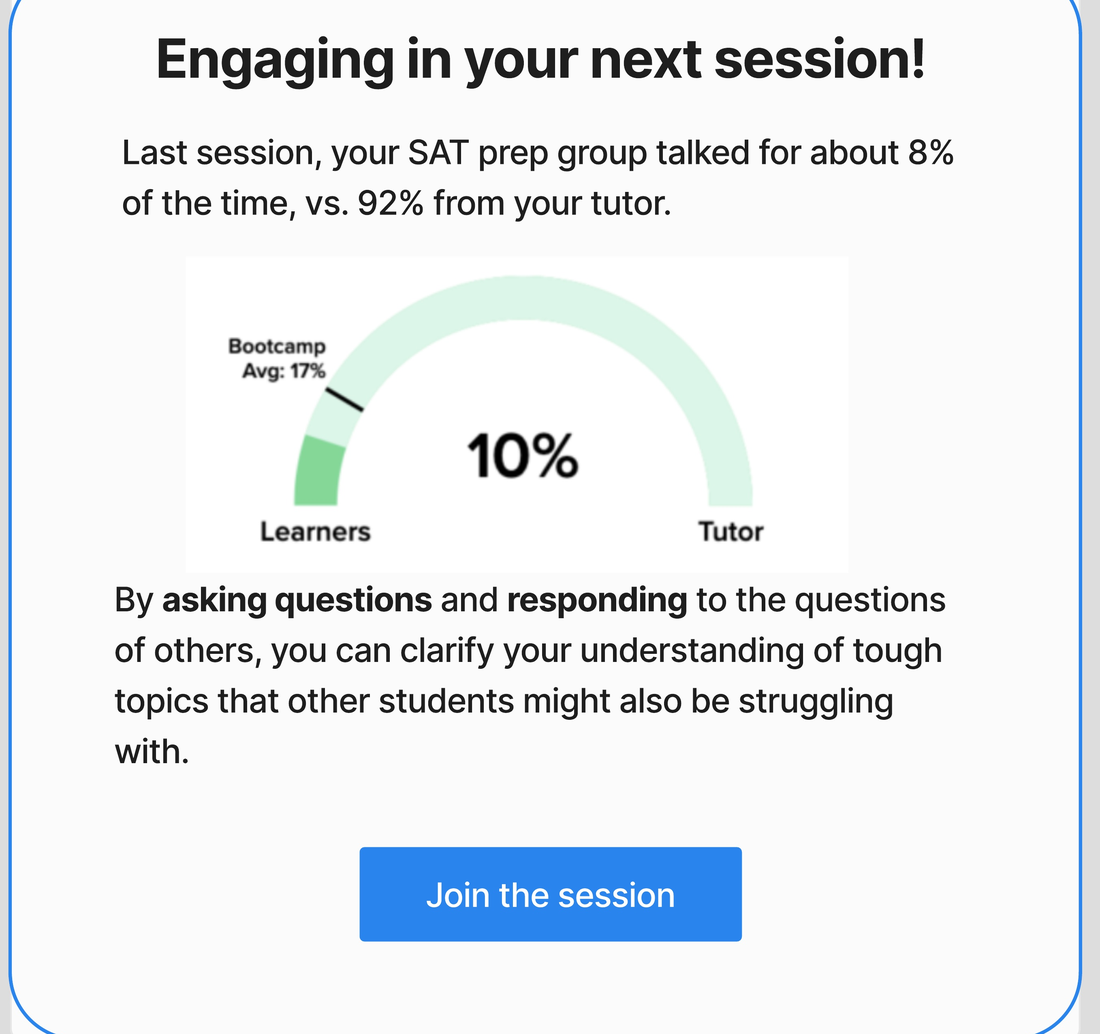}
\Description{Mockup of the learner feedback modal showing the share of talk by learners and the tutor.}\\[3pt]
{\small (b) Learner feedback}
\end{minipage}
\caption{The feedback shown to tutors and learners, from the platform's design
specification. (a) The primary tutor modal: summary statistics (1), highlighted moments from the tutor's own transcript followed by a fixed example from another tutor (2), and suggestions generated by GPT-4 Turbo (3). The reflection prompt (4) is omitted here for space. (b) The learner modal, shown in place of the usual join-session prompt. Figures are mockups with placeholder text and values; the two condition messages are given verbatim in Section~\ref{ssec:learner_feedback}.}
\label{fig:interface}
\end{figure*}

\subsection{Tutor feedback}
\label{ssec:tutor_feedback}

Tutor feedback was generated automatically from each session recording, usually within a few hours, and appeared as a modal on the platform before the tutor's next session. Tutors who had not opened it 24 hours before their next session also received an email reminder, and past feedback remained available on their profile.

The talk move in focus followed a fixed curriculum, applied to the feedback generated from each session: eliciting learner ideas from sessions 1 and 2, revoicing and building on learner ideas from sessions 3 and 4, and prompting for reasoning from sessions 6 and 7. A tutor therefore first saw feedback before their second session. No feedback was generated from session 5, which was dedicated to a full practice test, or from session 8.

Feedback came in two forms (Figure~\ref{fig:interface}). Feedback from the first session of each pair was a \emph{primary} modal with four components, which we refer to by number in Section~\ref{sec:rq3}. The first reports summary statistics: the tutor's talk time percentage and the proportion of learners who participated, each compared with the previous session and with the bootcamp average for the same session number. The second presents up to three moments from the tutor's own transcript illustrating the talk move in focus, followed by a fixed example drawn from another tutor, the same for every tutor in a given session. The third provides suggestions for the next session generated by GPT-4 Turbo from the tutor's transcript. The fourth invites the tutor to write a short reflection, which they could optionally share with other tutors. 

Feedback from the second session of each pair was a shorter \emph{follow-up} modal containing the first two components, the tutor's earlier reflection and a prompt to reflect on how it had gone. The follow-up modal contained no generated suggestions. Tutors therefore received generated suggestions roughly every other session rather than after every session.

\subsection{Learner feedback}
\label{ssec:learner_feedback}
In the two joint conditions, learners saw a feedback modal in place of the usual join-session modal when entering their next session. The modal reported the share of talk contributed by the learners and by the tutor in the previous session, alongside the bootcamp average, using the same component shown to tutors. Learners attending for the first time saw the usual join modal, and learners who had missed the previous session saw feedback from the last session they had attended.

The closing message varied by condition. Learners in the socially-oriented condition read: ``Remember: if you're struggling with something, some of your peers are probably struggling with the same thing! Sharing your questions and ideas can maximize everyone's learning during the session.'' Learners in the goal-oriented condition read: ``Asking questions and sharing your ideas will help you get personalized support on the specific areas you're struggling with, maximizing your score improvements!'' The rest of the modal was identical across the two conditions. The two framings, carried over from the pilot \cite{yun2024}, give learners different reasons to speak up. The socially-oriented message presents a learner's questions as useful to peers facing the same difficulty, drawing on evidence that a purpose beyond oneself can sustain effort on learning tasks \cite{yeager2014boring}. The goal-oriented message ties participation to the learner's own score, making its personal relevance explicit, as utility-value interventions do \cite{hulleman2009interest}. Comparing them tests whether learners in a small group respond more to a collective or a personal reason to participate.

Unlike tutors, learners received no email reminder and could not revisit past feedback, and every learner of a tutor in a joint condition saw the modal regardless of how much they had participated.

\subsection{Measures and generated suggestions}
\label{ssec:measures}

All discourse measures are computed from automated transcripts of the session recordings. Following \citet{yun2024}, we detect \emph{eliciting} by first filtering utterances with a fine-tuned question detection model and then classifying the results with an Electra-base model trained on the TalkMoves corpus, aggregating the pressing-for-reasoning and pressing-for-accuracy categories into a single positive class \cite{suresh2022talkmoves}. We also apply a measure of \emph{uptake} \cite{demszky2021uptake}, the extent to which a tutor builds on learner contributions. Two further measures serve as outcomes only and were never shown to tutors or learners: \emph{learner reasoning}, a classifier applied to learner utterances \cite{demszky2023nctetranscripts}, and the number of learner questions. We use hourly rates rather than raw counts, so that longer sessions do not mechanically produce higher counts.

We measure the balance of participation with two speech-based measures. \emph{Tutor talk time} is the tutor's share of speaking time, where consecutive utterances by the same speaker are treated as one turn. The \emph{proportion of learners speaking} is the number of attending learners with at least one spoken utterance divided by the number attending. Learners also participate by typing: they contribute 12\% of spoken words but 85\% of chat words, though the chat is under 1\% of all words in these sessions. We prioritize the time-based measure because it captures who holds the floor. Because the feedback displayed the tutor's share of words across speech and chat, and counted learners participating in either channel, we also report those two measures; the Supplementary Materials report all versions.\footnote{Supplementary Materials can be found here: \url{https://osf.io/aqy4u/overview?view_only=f3868c54875d4fb0b6f50e63fe9b4689}}

The generated suggestions were produced by GPT-4 Turbo from the tutor's transcript, using one prompt per weekly focus that supplied the platform's strategies for that focus and asked for a short summary and one to three specific points grounded in moments from the session. Before deployment, the prompts were tested on 80 outputs generated from existing transcripts, and tutor ratings from the pilot were used to improve them. The transcript moments shown alongside the suggestions were selected by the eliciting and uptake models and then filtered with GPT-4 Turbo to raise their precision.

Learner outcomes are the rating each learner supplies after a session on a three-point scale, a net promoter score collected every third session, and performance on practice tests, which we analyze both at the learner level and aggregated to the cohort.

\subsection{Sample}

Following the preregistered analysis plan we exclude session 1, which precedes any feedback and provides baseline covariates; session 5, after which no feedback was delivered; and recordings shorter than 30 minutes, which indicate a technical problem. This leaves 3,670 and 2,838 sessions in the two bootcamps. Our analytic sample (complete cases after listwise deletion on covariates) includes 3,170 sessions from 597 tutors for the first bootcamp and 2,483 sessions from 455 tutors for the second, and the pooled sample is 5,653 sessions from 899 tutors. The reduction is largely due to missing tutor gender, an optional profile field absent for 80 tutors in the first bootcamp and 57 in the second; Section~\ref{ssec:robustness} reports estimates without it. Learner-level models use 8,033 and 6,806 learner-sessions. Table~\ref{tab:desc} summarizes the sample and reports unadjusted means by condition.

\begin{table*}[t]
\centering\small
\caption{Tutors, sessions and unadjusted means by condition, before listwise deletion on model covariates. Talk-move measures are counts per hour; \emph{Engaged} is the proportion of attending learners who spoke at least once; \emph{Talk} is the tutor's share of speaking time, excluding the Zoom chat.}
\label{tab:desc}
\resizebox{.9\textwidth}{!}{%
\begin{tabular}{lrrrrrrr rrrrrrr}
\toprule
& \multicolumn{7}{c}{Bootcamp 1 (Apr--May)} & \multicolumn{7}{c}{Bootcamp 2 (May)} \\
\cmidrule(lr){2-8}\cmidrule(lr){9-15}
Condition & Tutors & Sess. & Elicit & Uptake & Reason & Engaged & Talk
          & Tutors & Sess. & Elicit & Uptake & Reason & Engaged & Talk \\
\midrule
Control             & 225 & 1190 & 17.7 & 6.1 & 1.63 & 0.81 & 0.917 & 174 & 948 & 16.1 & 5.5 & 1.58 & 0.79 & 0.914 \\
TutorFeedback       & 227 & 1225 & 19.9 & 6.7 & 1.86 & 0.83 & 0.912 & 174 & 954 & 18.7 & 6.3 & 1.84 & 0.80 & 0.918 \\
Tutor+GoalLearner   & 118 &  639 & 18.5 & 6.6 & 2.00 & 0.83 & 0.913 &  81 & 448 & 17.0 & 6.1 & 1.68 & 0.81 & 0.922 \\
Tutor+SocialLearner & 113 &  616 & 19.2 & 7.0 & 2.04 & 0.85 & 0.910 &  87 & 488 & 17.6 & 6.5 & 1.86 & 0.83 & 0.907 \\
\bottomrule
\end{tabular}
}
\end{table*}

\subsection{Regression analyses}
\label{ssec:regression_analysis}

We conducted a preregistered intention-to-treat analysis using ordinary least squares regression at the session level. We compare tutors by assigned condition regardless of whether they viewed their feedback, since compliance is likely correlated with tutor characteristics such as motivation, and since the effect of offering feedback is the quantity of interest for a platform deciding whether to adopt such feedback. The models are specified as $Y_{it} = \beta_1 T_i + \beta_2 X_i + \pi_t + \gamma_b + \varepsilon_{it}$, where $Y_{it}$ is a dependent variable for tutor $i$ in session $t$; $T$ indicates treatment status; $X$ is a vector of tutor-level and cohort-level covariates; $\pi_t$ are
session-number fixed effects; $\gamma_b$ are bootcamp fixed effects, included when the two bootcamps are pooled; and $\beta_1$ is the parameter of interest. For the first research question $T$ is a binary indicator for any treated condition. For the second we replace it with separate indicators for the tutor-only and joint conditions and report the difference between them; to compare the two framings we restrict the sample to the joint conditions. We cluster standard errors at the tutor level, including in the learner-level models, since learners within a tutor are not independent observations of treatment.

Covariates are measured before any feedback was delivered: the tutor's session 1 values for attendance, talk time, the proportion of learners speaking, and the rates of eliciting, uptake and learner reasoning; tutor gender and prior hosting experience on the platform; and the cohort's gender composition and SAT score band. Because we report several outcomes per family, we also report Benjamini--Hochberg corrected $p$-values within each family; the contrast between the joint and tutor-only conditions is reported uncorrected. Our primary specification pools both bootcamps, and we report each bootcamp separately throughout.

\label{ssec:deviations}\emph{Deviations from the preregistration.} We omit learner grade level, which the platform derives from date of birth and which is missing for 60.7\% of learners in the second bootcamp; retaining it would delete 49\% of that bootcamp's sessions, not at random. We compute baseline rates from the baseline session's own duration, cluster learner-level models on the tutor, the unit of randomization, and estimate the framing contrast within the joint conditions only. 

\subsection{Checking randomization}

We compared baseline characteristics across the four conditions, clustering at the tutor level. Two of twenty comparisons reach conventional significance, close to what chance produces: tutors' prior hosting experience in the first bootcamp ($p = .021$) and the baseline proportion of learners speaking in the second ($p = .024$). Both are included as covariates in all models. The Supplementary Materials report each condition's mean and a test of condition on each characteristic.

\subsection{Interviews}
\label{sec:method-interviews}

We interviewed 17 tutors after the first bootcamp: five from \textsc{TutorFeedback}, six from \textsc{Tutor+GoalLearner} and six from \textsc{Tutor+SocialLearner}. Tutors were sampled at random within condition, invited by email, and compensated with a \$15 gift card. Interviews lasted approximately 30 minutes, were conducted over Zoom by a member of the platform's staff, and followed a common protocol. Tutors were first asked general questions about the bootcamp; their own feedback was then shared on screen, and for each component they were asked how they felt about it, whether it was useful, whether they recalled acting on it, and whether acting on it was ever challenging.

We also interviewed 13 learners from the same bootcamp, sampled within condition: two from \textsc{TutorFeedback}, three from \textsc{Tutor+GoalLearner}, four from \textsc{Tutor+SocialLearner} and four from the control condition. All 13 answered questions about their own participation and about their group; the questions about the talk time feedback were administered to nine of them, including four of the seven learners in the joint conditions and all four control learners.

Transcripts were coded by question using an inductive scheme developed by the interviewer. No second coder was involved, so the counts we report describe the interviewed samples rather than supporting inference beyond them, and the interviewer being a member of the platform's staff may have encouraged favorable accounts. No control tutors were interviewed, so the tutor interviews describe how treated tutors experienced the feedback rather than a comparison between conditions; we use the control learners only as a check on delivery. All names are pseudonyms, and quotations are lightly edited to remove disfluencies.

\subsection{Use of AI tools}

The authors used Claude (Anthropic) to convert human-written Stata scripts into Python, to clean up human-written Python scripts, to tabulate
frequencies from human-assigned interview codes, to improve sentence structure, and to help extend list of human-provided references (all of which were then human-validated). Study design, the preregistered analysis plan, qualitative coding and all interpretation are the authors' own who take full responsibility for this paper's content. GPT-4 Turbo appears separately in this study as a component of the feedback.

\section{Results}

\begin{table*}[t]
\centering\small
\caption{Effect of automated feedback to tutors on instructional practice. Pooled model with bootcamp fixed effects ($n = 5{,}653$ sessions, 899 tutors); B1 and B2 are the same specification fit to each bootcamp separately ($n = 3{,}170$ / 597 and $2{,}483$ / 455). Standard errors clustered on the tutor. Effect ($\sigma$) is the coefficient in control standard deviations. BH: Benjamini--Hochberg corrected within the six-outcome family; the two
all-channel measures are alternative operationalizations of talk time and participation and are outside that family. $^{+}p<.10$, $^{*}p<.05$, $^{**}p<.01$.}
\label{tab:practice}
\begin{tabular}{lrrrrrrrr}
\toprule
& \multicolumn{6}{c}{Pooled} & \multicolumn{2}{c}{By bootcamp} \\
\cmidrule(lr){2-7}\cmidrule(lr){8-9}
Outcome & Coef. & (SE) & $p$ & BH $p$ & Ctrl.\ mean & Effect ($\sigma$) & B1 & B2 \\
\midrule
Eliciting / hr & 1.408$^{**}$ & (0.500) & 0.005 & 0.010 & 17.161 & 0.116 & 1.439$^{*}$ & 1.087 \\
Uptake / hr & 0.519$^{**}$ & (0.181) & 0.004 & 0.010 & 5.914 & 0.117 & 0.437$^{+}$ & 0.566$^{*}$ \\
Learner reasoning / hr & 0.238$^{**}$ & (0.074) & 0.001 & 0.008 & 1.637 & 0.123 & 0.255$^{*}$ & 0.196$^{+}$ \\
Learner questions / hr & 0.429$^{+}$ & (0.252) & 0.089 & 0.107 & 6.577 & 0.070 & 0.495 & 0.342 \\
Tutor talk time (speaking time) & $-$0.001 & (0.004) & 0.803 & 0.803 & 0.915 & $-$0.009 & $-$0.005 & 0.003 \\
Prop.\ learners speaking & 0.024$^{+}$ & (0.012) & 0.054 & 0.081 & 0.801 & 0.084 & 0.029$^{+}$ & 0.016 \\
\addlinespace
Tutor share of words, speech + chat & $-$0.004 & (0.004) & 0.304 & -- & 0.877 & $-$0.038 & $-$0.007 & $-$0.002 \\
Prop.\ learners participating, any channel & 0.013$^{+}$ & (0.007) & 0.055 & -- & 0.922 & 0.078 & 0.009 & 0.016 \\
\bottomrule
\end{tabular}
\end{table*}

\subsection{RQ1: does automated feedback to tutors change instruction and learner outcomes?}
\label{sec:rq1}

As Table~\ref{tab:practice} shows, feedback significantly increased all three discourse measures targeted by the curriculum. Pooling both bootcamps, treated tutors elicited learner ideas 1.41 more times per hour than control tutors, an 8\% increase over the control mean of 17.16 ($p = .005$). They built on learner contributions 0.52 more times per hour, a 9\% increase ($p = .004$), and their learners expressed reasoning 0.24 more times per hour, a 15\% increase ($p = .001$). All three effects correspond to approximately 0.12 control standard deviations and remain significant after correction for multiple comparisons. Two further measures point in the same direction without reaching conventional thresholds: the proportion of learners speaking increased by 2.4 percentage points ($p = .054$) and learner questions by 0.43 per hour ($p = .089$).

In contrast, we find no effect on tutor talk time ($-0.001$, $p = .803$), which averaged 91.5\% in the control group. The same holds for the statistic the feedback displayed, the tutor's share of words across speech and chat ($-0.004$, $p = .304$), and participation counted across both channels is marginal, like its spoken counterpart ($+0.013$, $p = .055$); estimates by channel are in the Supplementary Materials. This result differs from the pilot on the same platform, in which joint feedback increased tutor talk time \cite{yun2024}, and from 1:1 settings where the feedback decreased tutor talk time \cite{demszky2023mpowering,demszky2024talkmeter}. In Section~\ref{sec:rq3}, we seek to interpret this result by drawing on the interviews.

\begin{figure}[t]
\centering
\includegraphics[width=\linewidth]{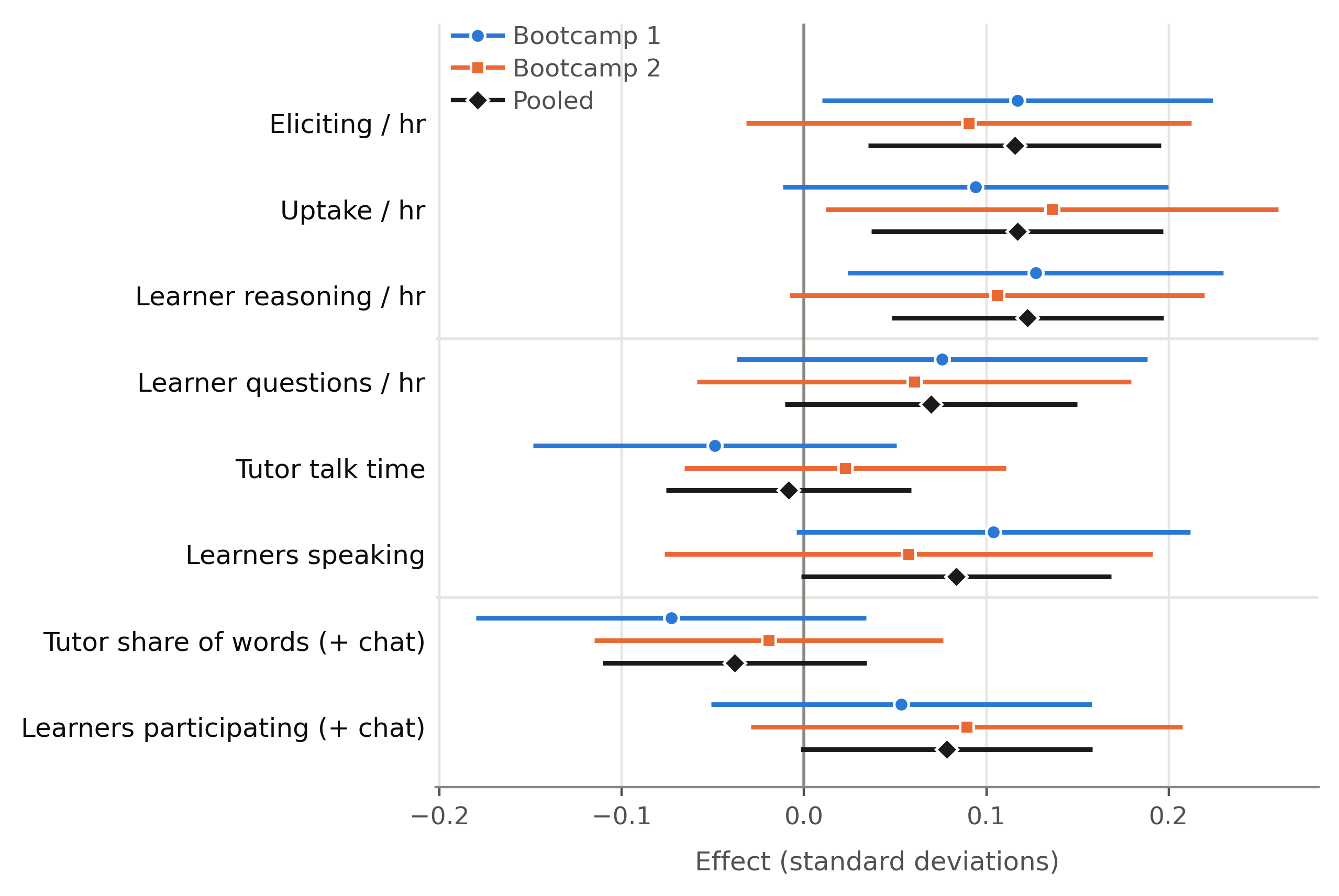}
\Description{Forest plot of standardized effects with 95 percent confidence intervals for each outcome, by bootcamp and pooled.}
\caption{Effects on instructional practice in standard deviation units, with 95\% confidence intervals, for each bootcamp and pooled. The three targeted discourse measures increase in both
bootcamps, while the balance of talk remains unchanged.}
\label{fig:forest}
\end{figure}

Estimated separately, no single-bootcamp effect survives correction for multiple comparisons. Before correction, the first bootcamp shows significant effects on eliciting ($p = .032$) and learner reasoning ($p = .016$), and the second on uptake ($p = .032$). The signs agree across bootcamps on five of the six practice measures; the exception, tutor talk time, is close to
zero in both (Figure~\ref{fig:forest}). The two bootcamps thus provide consistent, independent estimates of a modest effect.

Delivery was incomplete. Feedback was delivered for 77.9\% of treated learner-sessions in the first bootcamp (11,030 of 14,159) and 56.0\% in the second (6,868 of 12,275), where a delivery error withheld the feedback generated from session 1 from all but 20 of 342 treated tutors and the feedback from session 2 reached fewer than half of them. The weaker estimates in the second bootcamp are consistent with its lower delivery rate, but also with two other differences: only tutors in the first bootcamp were told in advance that they
would receive feedback, and 51 of the 174 control tutors in the second bootcamp had already received feedback, in the first bootcamp or the March pilot (Section~\ref{ssec:robustness}).

Turning to learner outcomes (Table~\ref{tab:outcomes}), we find no significant effects in the pooled models. At the cohort level,
attendance is $+0.03$ ($p = .71$), practice test performance $-0.005$ ($p = .30$), net promoter score $+0.008$ ($p = .91$) and rating $+0.007$ ($p = .66$). At the learner level the pooled estimates are likewise null, and the two bootcamps differ in sign on all three. Two single-bootcamp estimates do not survive correction and we do not interpret them: practice test performance in the second
bootcamp is negative ($-0.015$, $p = .038$ at the learner level), although practice tests are drawn from different forms and are not directly comparable across sessions, and learner session ratings are positive in the first ($+0.043$, $p = .049$). Over four weeks, then, the feedback changed observable discourse without changing attendance, measured learning or satisfaction.

\begin{table*}[t]
\centering\small
\caption{Effect of automated feedback to tutors on cohort and learner outcomes. Panel A is at the session level (3,858--5,653 sessions), Panel B at the learner-session level (5,840--14,839); both cluster on the tutor. Effect ($\sigma$) is the coefficient in control standard deviations. BH: Benjamini--Hochberg corrected within panel. $^{+}p<.10$, $^{*}p<.05$.}
\label{tab:outcomes}
\begin{tabular}{lrrrrrrrr}
\toprule
Outcome & Coef. & (SE) & $p$ & BH $p$ & Ctrl.\ mean & Effect ($\sigma$) & B1 & B2 \\
\midrule
\multicolumn{9}{l}{\emph{Panel A. Cohort outcomes}} \\
Attendance (learners) & 0.026 & (0.071) & 0.714 & 0.907 & 4.292 & 0.013 & 0.058 & $-$0.021 \\
Cohort practice test & $-$0.005 & (0.005) & 0.296 & 0.907 & 0.656 & $-$0.024 & 0.001 & $-$0.013$^{+}$ \\
Cohort NPS & 0.008 & (0.065) & 0.907 & 0.907 & 8.235 & 0.005 & 0.048 & $-$0.042 \\
Cohort rating & 0.007 & (0.015) & 0.655 & 0.907 & 2.383 & 0.020 & 0.023 & $-$0.011 \\
\midrule
\multicolumn{9}{l}{\emph{Panel B. Learner outcomes}} \\
Practice test & $-$0.007 & (0.005) & 0.171 & 0.392 & 0.652 & $-$0.027 & 0.001 & $-$0.015$^{*}$ \\
Session rating & 0.018 & (0.016) & 0.261 & 0.392 & 2.354 & 0.034 & 0.043$^{*}$ & $-$0.007 \\
NPS & $-$0.005 & (0.073) & 0.949 & 0.949 & 8.290 & $-$0.002 & 0.127 & $-$0.161 \\
\bottomrule
\end{tabular}
\end{table*}

\subsection{Robustness}
\label{ssec:robustness}

Full estimates for the three checks below are in the Supplementary Materials.

\emph{Prior exposure.} Our bootcamps and the March pilot drew on the same pool of tutors: 228 of our 1{,}030 tutors had also tutored in the pilot \cite{yun2024}, and 169 taught in both of our bootcamps. As a result, 38 of 225 control tutors in the first bootcamp and 51 of 174 in the second had already received feedback, although the dosage of the pilot feedback was lower and did not cover prompting for reasoning. Restricting to tutors with no prior feedback in either arm (3{,}739 sessions, 697 tutors) yields effects of 11.2\% for eliciting, 8.0\% for uptake and 17.1\% for learner reasoning, all significant, and raises the second-bootcamp estimate for eliciting from 1.09 ($p = .147$) to 1.98 per hour ($p = .034$). Tutor talk time remains null pooled; in that restricted second-bootcamp sample it is 1.4 percentage points higher among treated tutors ($p = .049$, BH $p = .147$), which we do not interpret, though its direction matches the pilot \cite{yun2024}.

\emph{Delivery.} Instrumenting delivery with assigned condition (first-stage $F > 4{,}000$) scales the estimates up by roughly a sixth, to 9.6\% for eliciting, 10.3\% for uptake and 17.1\% for learner reasoning, with talk time null. With a single binary instrument this rescales the intention-to-treat estimate and leaves $p$-values unchanged, so we treat it as a check on magnitude, one that assumes the advance email did not itself change behavior.

\emph{Covariates.} Dropping the two gender covariates recovers 810 sessions (6{,}463 sessions, 1{,}021 tutors) and makes all three targeted estimates larger and more precise (all $p < .001$); talk time remains null. Learner questions then becomes significant after correction (0.54, BH $p = .042$). Because it does so only without the gender covariates, and neither it nor the proportion of learners speaking survives the prior-exposure restriction, we treat both as suggestive.

\subsection{RQ2: does learner feedback change outcomes further?}
\label{sec:rq2}

Table~\ref{tab:learner} reports the difference between the joint conditions and \textsc{TutorFeedback} alone. We find no effect of learner feedback on any measure of instructional practice, in the pooled model or in either bootcamp, including the measures that credit learners for typing in the chat: the tutor's share of words across both channels is $-0.004$ ($p = .42$) and the proportion of learners participating in any channel $+0.008$ ($p = .24$).

\begin{table*}[t]
\centering\small
\caption{Learner feedback above tutor feedback alone: the contrast between the two joint conditions and \textsc{TutorFeedback}. Effect ($\sigma$) is the coefficient in control standard deviations. Sample sizes as in Tables~\ref{tab:practice} and~\ref{tab:outcomes}. $p$-values are \emph{uncorrected}; correcting across the four satisfaction measures would leave none below .05. $^{+}p<.10$, $^{*}p<.05$.}
\label{tab:learner}
\begin{tabular}{lrrrrrrr}
\toprule
Outcome & Coef. & (SE) & $p$ & Ctrl.\ mean & Effect ($\sigma$) & B1 & B2 \\
\midrule
\multicolumn{8}{l}{\emph{Panel A. Instructional practice}} \\
Eliciting / hr & $-$0.731 & (0.607) & 0.228 & 17.161 & $-$0.060 & $-$1.009 & $-$0.069 \\
Uptake / hr & 0.232 & (0.211) & 0.272 & 5.914 & 0.052 & 0.199 & 0.320 \\
Learner reasoning / hr & 0.067 & (0.091) & 0.466 & 1.637 & 0.034 & 0.210$^{+}$ & $-$0.115 \\
Learner questions / hr & $-$0.050 & (0.314) & 0.874 & 6.577 & $-$0.008 & $-$0.140 & 0.123 \\
Tutor talk time (speaking time) & $-$0.000 & (0.004) & 0.952 & 0.915 & $-$0.002 & $-$0.000 & $-$0.000 \\
Prop.\ learners speaking & 0.014 & (0.012) & 0.256 & 0.801 & 0.049 & 0.019 & 0.008 \\
Tutor share of words, speech + chat & $-$0.004 & (0.005) & 0.418 & 0.877 & $-$0.034 & $-$0.007 & 0.002 \\
Prop.\ learners participating, any channel & 0.008 & (0.007) & 0.242 & 0.922 & 0.049 & 0.015 & 0.002 \\
\midrule
\multicolumn{8}{l}{\emph{Panel B. Cohort outcomes}} \\
Attendance (learners) & 0.034 & (0.079) & 0.661 & 4.292 & 0.018 & $-$0.024 & 0.125 \\
Cohort practice test & $-$0.003 & (0.005) & 0.581 & 0.656 & $-$0.014 & $-$0.005 & 0.002 \\
Cohort NPS & 0.153$^{*}$ & (0.074) & 0.039 & 8.235 & 0.095 & 0.196$^{+}$ & 0.103 \\
Cohort rating & 0.028$^{+}$ & (0.016) & 0.086 & 2.383 & 0.081 & 0.021 & 0.034 \\
\midrule
\multicolumn{8}{l}{\emph{Panel C. Learner outcomes}} \\
Practice test & $-$0.002 & (0.005) & 0.708 & 0.652 & $-$0.008 & 0.002 & $-$0.006 \\
Session rating & 0.033$^{+}$ & (0.018) & 0.073 & 2.354 & 0.063 & 0.019 & 0.053$^{+}$ \\
NPS & 0.180$^{*}$ & (0.082) & 0.028 & 8.290 & 0.095 & 0.152 & 0.259$^{*}$ \\
\bottomrule
\end{tabular}
\end{table*}

We do observe small differences in how learners rated their experience: learner net promoter score is $+0.18$ ($p = .028$), cohort net promoter score $+0.15$ ($p = .039$), and learner session rating $+0.033$ ($p = .073$). These $p$-values are
uncorrected, none would remain significant after correcting across the four satisfaction measures, and the net promoter result is carried by the second bootcamp. Thus, while learner feedback marginally increased how learners rated their sessions, it did not observably change discourse in the sessions.

Comparing the two framings within the joint conditions, no outcome differs significantly; the smallest $p$-value across ten outcomes is .47, estimated on 336 tutors and 1,917 sessions. This is our least well powered comparison, and its confidence intervals include effects as large as those we detect elsewhere, so it does not establish that framing has no effect. Section~\ref{sec:rq3} draws on learner interviews to interpret these results.

\subsection{RQ3: how did tutors and learners perceive the feedback, and what were the barriers?}
\label{sec:rq3}

All 17 tutors we interviewed reported reading their feedback. Estimates of how long they spent on it per session ranged from under a minute to fifteen minutes, clustering at two to five, and five tutors described spending less time as the bootcamp progressed. Stefan: ``as the boot camp went on, I spent less and less time looking at the feedback. I would just skim over it.'' Three themes emerged from the tutor interviews; code frequencies by condition are in the Supplementary Materials.

\emph{Tutors acted on the summary statistics by trying to redistribute talk.} Twelve of 17 tutors said the statistics encouraged them and three that they were discouraging. Fourteen of 17 recalled acting on this component, and they consistently described trying to talk less and draw learners in. Anaya said, ``one of the first days is when I realized that I might talk too much. And so the next tutoring session, I kind of made my students talk more and explain how
they went through things.''

\emph{The generated suggestions led tutors to adopt specific discourse moves.} Fourteen of 17 tutors found the third component useful, most often describing it as constructive criticism (9 of 17), and 11 recalled acting on it. Three tutors independently described the same change. Skylar said, ``instead of saying, do we have any questions? I can say what questions we have instead\ldots\ every time I was going to ask questions, I reminded myself to say, what are our questions.'' Anaya: ``they basically told me instead of asking, does anyone have any questions, to
say what questions do you have. I started doing that.'' This substitution was specified in the prompt used to generate the feedback, which instructed the model to ``normalize learners asking questions by asking for questions using language that implies that questions are expected. E.g., instead of `any questions?' ask `what questions do you have?'\,'' The instruction, the generated feedback and the behavior tutors report therefore correspond directly. Other tutors described changes corresponding to uptake. Stefan said, ``it told me I could do a better job of reinforcing what my learners are saying\ldots\ And I definitely did make a note to do that better in the next session.''

\emph{The main barrier to acting on the feedback was learners who did not respond.} Asked what was most challenging about tutoring, 10 of 17 tutors named getting learners to engage, by a wide margin the most common answer. Ten of 17 said that acting on the summary statistics had at some point been difficult, and 8 attributed the difficulty to learners who would not participate. Asha said, ``sometimes if I ask the learners a question, they won't respond, or none of them will say anything. And I don't usually like cold calling.'' Lila connected the barrier directly to talk time: ``I still talk too much because I need to explain the problems, and sometimes if they don't ask questions\ldots\ I'll just try to re-explain what I've explained.'' Bryce distinguished between attempting the change and achieving it: ``Maybe the actual part, increasing participation, is hard, but the act of trying to increase it is not.'' Two tutors questioned the accuracy of the statistics; Daniel said the feedback ``would be more careful if it could read the chat.''

Finally, tutors differed sharply in how useful they found the four components. The second component, which highlighted moments from the transcript, was the least useful: 10 of 17 named it least helpful, 13 of 17 never acted on it, and 12 said they had not paid attention to it. Every tutor who called it unhelpful described it as redundant. Kieran said, ``the least helpful section was probably section two, since section three usually covers the information.'' Vaani objected to the peer examples, which were identical for every tutor in a given session: ``I want
to focus on myself as a tutor rather than look at snippets of other tutors.'' The summary statistics and the generated suggestions were named most helpful equally often (8 of 17 each), and were the components tutors described acting on.

\emph{Learners understood the learner feedback but could not act on it.} Learner interviews help explain why learner feedback did not change practice (Section~\ref{sec:rq2}). Three of the four control learners did not recall seeing the feedback, while learners in the joint conditions recalled it readily; one control learner reported having seen it once, consistent with the eight control sessions for which feedback was recorded as delivered. Delivery therefore appears to have worked largely as intended. 
The accounts below come from the four learners in joint conditions who were asked about the feedback, and from the participation questions answered by all thirteen.

First, the quantity they were shown was a property of the group, not of themselves. Noah reported checking the feedback ``all the time, unfortunately'', and then explained why it changed little: ``even when I try to talk a little bit more, it would still say, hey, y'all are talking a lot less. So I was like, you know what, just whatever.'' It was only ``slightly useful\ldots\ since it's dependent on all learners.'' He also attributed the imbalance to his tutor rather than to himself.

Second, some learners did not want to talk more. Noah again: ``I want to hear more what the tutor has to say versus what anyone else has to say\ldots\ it didn't feel like my place to try to up that bar when I don't need to talk about anything.''

Third, learners disputed the measure. Livia said the feedback ``really wasn't an accurate reflection of how we were engaging because it was talking about how much we were talking out loud. A couple of times it said like 1\% engagement from us when we were engaged. We just were talking in the chat instead of out loud.'' Aurelia, shown her own feedback during the interview, said simply, ``I don't believe this data.'' The statistic did count chat words, but because typed messages are short and the chat is under 1\% of all words, learners who participated by typing barely changed it. Tutors raised the same objection.

Fourth, and most relevant to the framing comparison, the socially-oriented message asked learners to participate for the benefit of their peers where peer-to-peer interaction barely existed. Every learner we interviewed described interaction as predominantly between tutor and learner, with estimates from 90 to 98 percent. Aiden: ``in literally all four camps I did, there was never learner to learner. That never happened.'' Marin identified a platform constraint: ``private messaging between other members of the Zoom that weren't the tutor was off, so\ldots\ we never really had the chance to interact with each other.''

\section{Discussion}

\subsection{The impact of tutor feedback}

Automated feedback increased tutors' eliciting and uptake, and their learners' reasoning, by 8--15\% across two bootcamps, and left the balance of talk unchanged. The interviews suggest why. Eliciting, uptake and prompting for reasoning are moves a tutor can make on their own initiative in the next session, and tutors described doing so, several naming the exact wording they changed. The share of talk is jointly produced, tutors repeatedly described learners who did not want to speak more. This finding corroborates prior research on group participation displays, which found the same asymmetry of dominant speakers talking less but quieter members not talking more \cite{dimicco2004,bachour2010}.

An alternative explanation is that the feedback reduced urgency for some tutors: those already above the bootcamp average on learner participation saw largely favorable statistics, and the encouraging tone of the generated suggestions may have reassured the rest. In an exploratory analysis, however, the effect on talk time did not vary with tutors' baseline talk time (interaction $p = .33$), though that test has limited power.

For the design of discourse analytics, this suggests distinguishing measures of what an educator does from measures of what the group does. Both can appear equally actionable on a dashboard, but only the first maps onto a specific action the educator can take. Moving the second may require support beyond feedback to one person, such as coaching that uses the same analytics \cite{sawaya2025,malamut2026facilitating}, actionable group-level suggestions, or session structures that give learners more occasions to speak, such as brief pair or small-group work within the session, which the platform's Zoom configuration currently limits (Section~\ref{sec:rq3}). Analytics that model what learners contribute to one another, rather than their share of talk, could then support those structures \cite{chandler2024collab}.

\subsection{The (lack of) impact of learner feedback}

If unresponsive learners are the binding constraint, learner-facing feedback would appear to be the right lever, yet it had no effect on practice. The reasons learners gave (Section~\ref{sec:rq3}) rest on four learners who received the feedback, so we offer them as hypotheses. The first two, that the statistic described the group rather than the learner and that some learners preferred to listen, align with evidence that learners rarely use participation supports \cite{strauss2021} and that the paths from participation feedback to action vary across learners \cite{ochoa2024}. In 1:1 tutoring, where the Talk Meter was effective \cite{demszky2024talkmeter}, the learner's behavior determines the statistic entirely, but in a group of up to ten it does not. Analytics that describe each learner's own participation are now possible to obtain automatically \cite{zhou2024transparent} and have made collaborative dialogue more equitable \cite{hu2024}, which points to learner-level feedback as the more promising design. 

The third, that learners doubted the measure, reflects how the statistic weighted channels: typed messages counted, but as words, so they barely registered. Learners' uptake of feedback depends on how they appraise it \cite{carless2018}, and quieter participants in earlier work likewise judged participation displays less accurate \cite{dimicco2004}. Crediting each learner's typed and spoken contributions comparably, for instance by counting turns rather than words, is a solvable problem and may be a precondition for learner-facing participation feedback.

A fourth reason is timing and dose. The Talk Meter appeared during the session and gave tutor and student a shared, visible goal. Our learner feedback appeared once, as learners joined the next session, and could not be revisited, whereas tutors could reread past feedback and saw suggestions tied to their own transcript. Learners thus received a thinner intervention than tutors did. Finally, the pro-social framing presupposed peers who interact, and learners in every condition described learner-to-learner interaction as rare; testing that framing fairly would require a session structure in which peers can act on each other's contributions.
\subsection{Design implications}

Three findings bear on the design of similar systems. First, tutors acted on components that named a concrete next step. Teachers can reflect on accurate feedback
without changing their practice \cite{kelly2024}; here, the summary statistics and generated suggestions were read and acted on, while the retrieved excerpts, including a peer example identical for every tutor, were ignored as redundant. Second, engagement declined over the bootcamp, which suggests front-loading the most substantive feedback and keeping each component brief. Third, delivery is a first-order problem: roughly a third of treated learner-sessions received no feedback, and a system's effect is bounded by how reliably it reaches its users.

These findings have already informed practice on the platform. Drawing on the tutor interviews, Schoolhouse.world redesigned its automated feedback for Spring 2025. It expanded the talk time component, the one tutors most often acted on, into a richer visualization of the tutor's talk time. It removed the highlighted transcript moments, which most tutors had not acted on, and the count of learners who participated, which tutors found redundant. It also shortened the generated suggestions, which tutors valued but found long, to one- or two-sentence notes on specific aspects of the session. The platform also revised its tutor training to give tutors more concrete strategies for building relationships with learners and adapting to them in real time, which responds to the barrier tutors described most often. Whether these changes shift the balance of talk remains to be evaluated.

\section{Limitations}

Our estimates are intention-to-treat under incomplete delivery, and some control tutors had received feedback before the trial; both likely attenuate them (Section~\ref{ssec:robustness}). We lack log data on whether or when tutors opened their feedback, so what we know about how they engaged with it comes from interviews and from the reflections they wrote, which declined over each bootcamp. Our outcome window is four weeks, and without scores from the subsequent SAT administration we cannot say whether the changes in discourse persist or affect learning.

The classifiers detect the presence of a move, not its quality or its fit to the moment, and only the eliciting and uptake models were validated on transcripts from this platform. Tutors also reported transcription errors. 

The interviews come from one bootcamp, were conducted by a member of the platform's staff, and were coded by a single coder. No control tutors were interviewed, and our account of learners' responses rests on few
interviews. The comparison between the two motivational framings is underpowered.

Finally, our sample consists of volunteer, mostly high school tutors on a single platform, preparing learners for a high-stakes test in sessions where learners rarely spoke. The suggestions came from GPT-4 Turbo, and newer models may produce different feedback. We do not know how these results generalize to paid tutors, classroom teachers, lower-stakes settings or other session formats.

\section{Conclusion}

In a randomized controlled trial across two SAT bootcamps with 1{,}030 volunteer tutors, automated feedback on session transcripts increased tutors' eliciting and uptake of learner ideas, and learners' reasoning, by 8 to 15 percent, while tutors continued to do about 90 percent of the talking. Tutors adopted the specific moves the feedback suggested, and the main barrier they described was learners who did not respond. Participation feedback delivered to learners did not change discourse further.

Automated feedback thus changed what tutors could do on their own initiative, while the balance of talk, which depends on the whole group, did not change. Shifting it may require changing the form of feedback, and session structures that give learners more occasions to speak with the tutor and with each other.

\begin{acks}
We thank all the tutors and learners for their participation in the study and the interviews. We are grateful to Cassy McKee and other members of the Schoolhouse.world team for their support.

\textbf{Author contributions (CRediT).}
Conceptualization, Supervision, Writing -- original draft: D.D.;
Data curation: Y.H., J.Y., J.H.;
Formal analysis: D.D., M.O., M.K.;
Funding acquisition: D.D., M.W., D.B.;
Investigation: D.D., M.O., M.K.;
Project administration: M.O.;
Software: A.I., A.R., J.H.;
Writing -- review \& editing: M.O., M.W.;

\textbf{Declaration of interests.} M.O., M.K., A.I., M.W., J.H., A.R.\ and D.B.\ are or were affiliated with Schoolhouse.world, the nonprofit platform on which the study was conducted. Schoolhouse.world had no financial interest in the results of the study. The analysis plan was preregistered, and all quantitative analyses were conducted by D.D., who is not affiliated with Schoolhouse.world. The interviews were conducted by a member of the platform's staff (Section~\ref{sec:method-interviews}).

\end{acks}

\bibliographystyle{ACM-Reference-Format}
\bibliography{references}

\end{document}